\documentclass[10pt,twocolumn,letterpaper]{article}

\usepackage{cvpr}              

\usepackage{graphicx}
\usepackage{amsmath}
\usepackage{amssymb}
\usepackage{booktabs}

\usepackage[pagebackref,breaklinks,colorlinks]{hyperref}

\usepackage[capitalize]{cleveref}
\crefname{section}{Sec.}{Secs.}
\Crefname{section}{Section}{Sections}
\Crefname{table}{Table}{Tables}
\crefname{table}{Tab.}{Tabs.}

\def\cvprPaperID{1234} 
\def\confName{CVPR}
\def\confYear{2023}

\begin{document}

\title{iQuery: Instruments as Queries for Audio-Visual Sound Separation}

\author{Jiaben Chen$^{1}$, Renrui Zhang$^{2,3}$, Dongze Lian$^{4}$, Jiaqi Yang$^{5}$, Ziyao Zeng$^{5}$, \vspace{0.3cm} Jianbo Shi$^{6}$ \\
  $^1$UC San Diego \quad 
  $^2$Shanghai AI Laboratory \quad\\
  $^3$The Chinese University of Hong Kong \quad
  $^4$National University of Singapore\\
  $^5$ShanghaiTech University \quad
  $^6$University of Pennsylvania\vspace{0.2cm}\\
\texttt{jic088@ucsd.edu}, \quad \texttt{zhangrenrui@pjlab.org.cn}, \quad \texttt{dongze@nus.edu.sg}, \\ \texttt{\{yangjq, zengzy\}@shanghaitech.edu.cn}, \quad \texttt{jshi@seas.upenn.edu}
}

\maketitle

\begin{abstract}

Current audio-visual separation methods share a standard architecture design where an audio encoder-decoder network is fused with visual encoding features at the encoder bottleneck. This design confounds the learning of multi-modal feature encoding with robust sound decoding for audio separation. To generalize to a new instrument: one must finetune the entire visual and audio network for all musical instruments.   We re-formulate visual-sound separation task and propose Instrument as Query (iQuery) with a flexible query expansion mechanism. Our approach ensures cross-modal consistency and cross-instrument disentanglement. We utilize ``visually named" queries to initiate the learning of audio queries and use cross-modal attention to remove potential sound source interference at the estimated waveforms. To generalize to a new instrument or event class, drawing inspiration from the text-prompt design, we insert an additional query as an audio prompt while freezing the attention mechanism. Experimental results on three benchmarks demonstrate that our iQuery improves audio-visual sound source separation performance. Code will be available at \url{https://github.com/JiabenChen/iQuery}.


   
\end{abstract}

\section{Introduction \label{1}}

Humans use multi-modal perception to understand complex activities.  To mimic this skill, researchers have studied audio-visual learning \cite{hu2019deep,afouras2020self,cheng2020look} by exploiting the synchronization and correlation between auditory and visual information.  In this paper, we focus on the sound source separation task, where we aim to identify and separate different sound components within a given sound mixture~\cite{roweis2000one, virtanen2007monaural}. Following the ``Mix-and-Separate'' framework \cite{huang2015joint,hershey2016deep,yu2017permutation}, we learn to separate sounds by mixing multiple audio signals to generate an artificially complex auditory representation and then use it as a self-supervised task to separate individual sounds from the mixture.  The works \cite{owens2018audio,gao2019co,zhao2018sound} showed visually-guided sound separation is achievable by leveraging visual information of the sound source.




Prevalent architectures take a paradigm of a visual-conditioned encoder-decoder architecture \cite{gao2019co,zhao2019sound,gan2020music,rahman2021tribert}, where encoded features from audio and visual modalities are fused at the bottleneck for decoding to yield separated spectrogram masks. However, it is noticed that this design often creates a ``muddy" sound and ``cross-talk'' that leaks from one instrument to another.
To create a clean sound separation, one would like the audio-video encoders to be (1) self-consistent within the music instrument and (2) contrasting across. One approach \cite{gao2021visualvoice} added critic functions explicitly to enforce these properties. Another method~\cite{zhu2022visually} used a two-step process with the second motion-conditioned generation process to filter out unwanted cross-talks. We call these approaches decoder-centric.

Most recent works focus on addressing the ``muddy" and ``cross-talk" issue by improving fine details of audio-visual feature extraction: for example, adding human movement encoding as in \cite{zhao2019sound,gan2020music,zhu2022visually}, or cross-modality representations \cite{rahman2021tribert} via self-supervised learning.  
Once the feature representations are learned, the standard encoder-decoder FCN style segmentation is used as an afterthought. We consider these methods feature-centric. The standard designs have two limitations.  First, it is hard to balance decoder-centric and feature-centric approaches that enforce a common goal of cross-modality consistency and cross-instrument contrast. Second, to learn a new musical instrument, one has to retrain the entire network via self-supervision.  

 
\begin{figure*}[ht]
\begin{center}
    \includegraphics[width=0.95\linewidth]{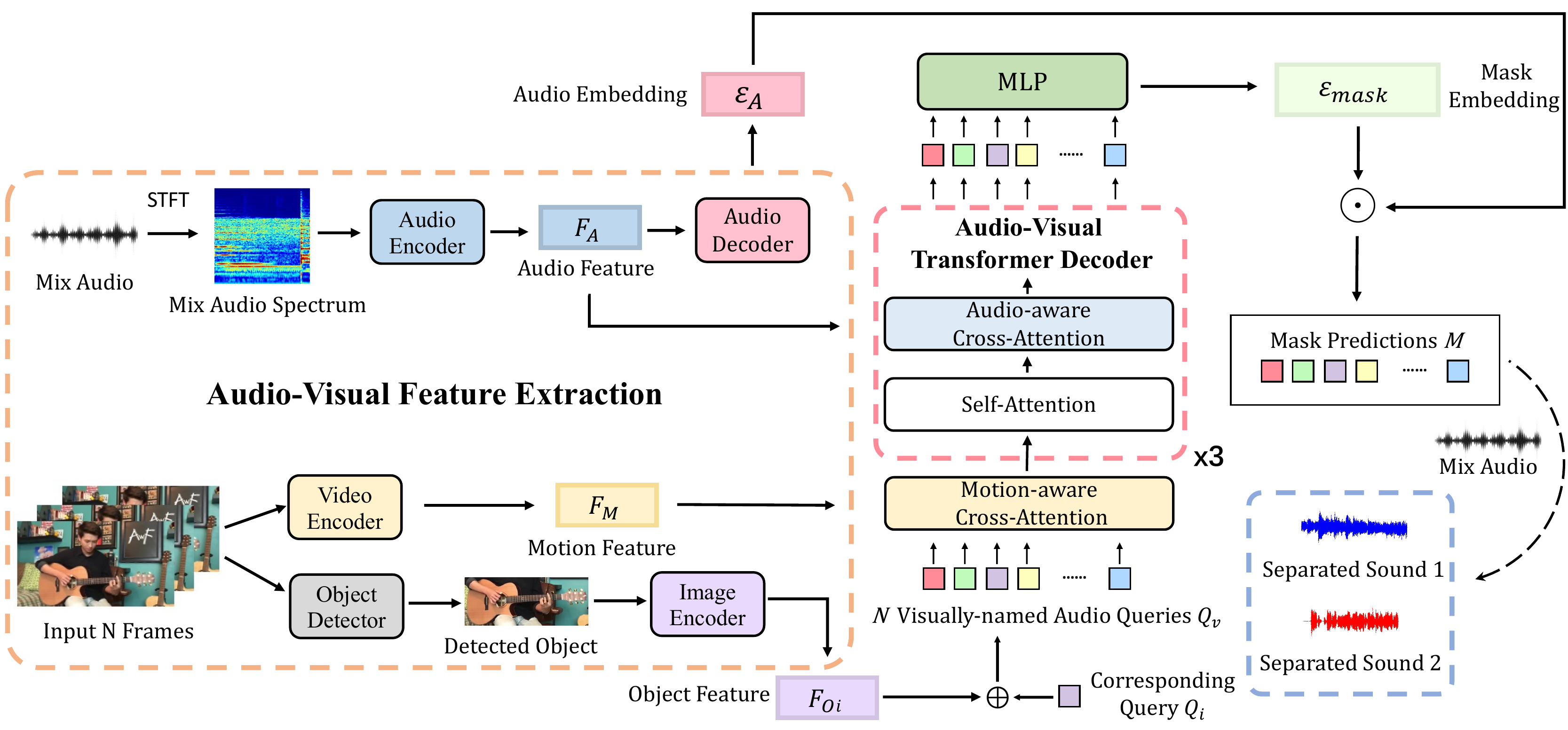}
\end{center}
\vspace{-4mm}
\caption{ \textbf{Pipeline of iQuery.} Our system takes as input an audio mixture and their corresponding video frames and disentangles separated sound sources for each video. Our pipeline consists of two main modules: an Audio-Visual Feature Extraction module which extracts audio, object, and motion features through three corresponding encoders, and an Audio-Visual Transformer module for sound separation. The query-based sound separation Transformer has three key components: \textbf{1)} ``visually-named" audio queries are initialized by extracted object features, \textbf{2)} cross-attention between the audio queries with static image features, dynamic motion features and audio features, \textbf{3)} self-attention between the learned audio queries to ensure cross-instrument contrast.}
\label{fig:method}
\end{figure*}
To tackle these limitations, we propose a query-based sound separation framework, iQuery. We recast this problem from a query-based transformer segmentation view, where each query learns to segment one instrument, similar to visual segmentation \cite{strudel2021segmenter,xie2021segformer,cheng2021per,cheng2021masked}.  We treat each audio query as a learnable prototype that parametrically models one sound class.  We fuse visual modality with audio by ``visually naming" the audio query:  using object detection to assign visual features to the corresponding audio query.  Within the Transformer decoder, the visually initialized queries interact with the audio features through cross-attention, thus ensuring cross-modality consistency.   Self-attention across the audio queries for different instruments implements a soft version of the cross-instrument contrast objective.  With this design, we unify the feature-centric with the decoder-centric approach.

How do we achieve generalizability?  Motivated by recent success in fine-tuning domain transfer with the text-prompt \cite{gao2020making} and visual-prompt designs \cite{bahng2022visual,jia2022visual,zhang2022neural,Lian_2022_SSF}, we adaptively insert the additional queries as audio prompts to accommodate new instruments.   With the audio-prompt design, we freeze most of the transformer network parameters and only fine-tune the newly added query embedding layer.  
We conjecture that the learned prototype queries are instrument-dependent, while the cross/self-attention mechanism in the transformer is instrument-independent.

Our main contributions are:
\begin{itemize}
    \item To the best of our knowledge, we are the first to study the audio-visual sound separation problem from a tunable query view to disentangle different sound sources explicitly through learnable audio prototypes in a Mask Transformer architecture.
    \item  To generalize to a new sound class, we design an audio prompt for fine-tuning with most of the transformer architecture frozen.
    \item Extensive experiments and ablations verify the effectiveness of our core designs for disentanglement, demonstrating performance gain for audio-visual sound source separation on three benchmarks.
\end{itemize}






\section{Related work \label{2}}
\begin{figure*}[ht]
\begin{center}
    \includegraphics[width=0.8\linewidth]{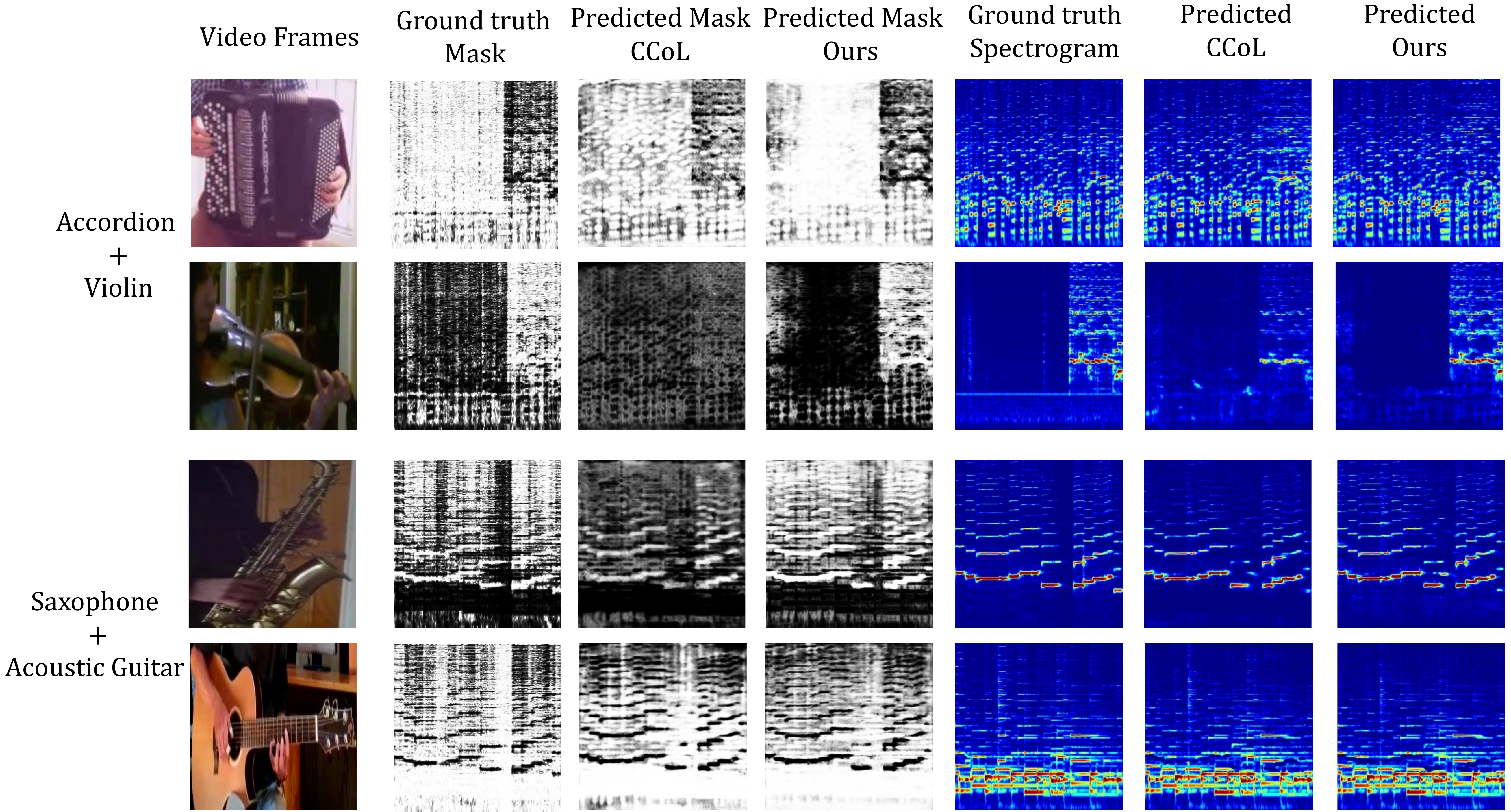}
\end{center}
\caption{\textbf{Qualitative results on \emph{MUSIC} test set.} The first column shows the mixed video frames, the second to the fourth columns compare our predicted spectrogram masks against masks yielded by state-of-the-art algorithm \cite{tian2021cyclic} and ground truth masks, and the fifth to the seventh columns visualize separated spectrograms. \cite{tian2021cyclic} produces blurry masks and contains unseparated components from another sound source, while our system successfully generates accurate mask and clean spectrograms as the ground truth.}
\label{fig:qualitative}
\end{figure*}
\noindent\textbf{Audio-visual Sound Source Separation.} Recent years have witnessed promising results of audio-visual multi-modality joint learning \cite{tian2020unified,zellers2022merlot,mercea2022temporal,wu2022wav2clip,shi2022learning} in domains like audio-visual sound source localization \cite{kidron2005pixels,zhou2016learning,arandjelovic2018objects,arandjelovic2017look,senocak2018learning,qian2020multiple,chen2021localizing,song2022self}, audio-visual event localization \cite{tian2018audio,wu2019dual,zhou2021positive,xia2022cross} and sound synthesis from videos \cite{owens2016visually,zhou2018visual,morgado2018self,gao20192,xu2021visually}.
Sound source separation, a challenging classical problem, has been researched extensively in the audio signal processing area \cite{fevotte2009nonnegative, cichocki2009nonnegative,chandna2017monoaural,lee2019audio,kilgour2022text}. A well-known example is the cocktail party problem \cite{haykin2005cocktail,mcdermott2009cocktail} in speech domain \cite{afouras2018conversation,ephrat2018looking}. Works have been proposed recently for tasks like speech separation\cite{afouras2019my,gao2021visualvoice,lee2021looking,truong2021right,montesinos2022vovit}, active sound separation \cite{majumder2021move2hear,majumder2022active} and on-screen sound separation \cite{owens2018audio,gao20192,tzinis2020into,tzinis2022audioscopev2}. 
Our work focuses on audio-visual sound separation. Recent audio-visual sound separation methods could be classified generally into two categories: feature-centric and decoder-centric as discussed in Sec. \ref{1}. Feature-centric methods exploit various ways for visual feature extraction selection to aid this multi-modality task. Some works consider frame-based appearance features (static frame features \cite{zhao2018sound,gao2018learning,xu2019recursive} or detected object regions \cite{gao2019co,tian2021cyclic}) for extracting visual semantic cues (\emph{e.g.}, instrument categories) to guide sound separation. \cite{chatterjee2021visual,chatterjee2022learning} adds embeddings from an audio-visual scene graph at the U-Net bottleneck to model the visual context of sound sources. Based on the assessment that motion signals could more tightly couple the moving sounding object with corresponding variations of sounds,  recent approaches focus on including motion information into the pipeline (\emph{e.g.}, optical flow \cite{zhao2019sound}, and human pose \cite{gan2020music,rahman2021tribert}). Based on this, \cite{zhou2022sepfusion} proposes a framework to search for the optimal fusion strategy for multi-modal features. Decoder-centric methods explore prevention of ``cross-talk'' between the audio sources in the decoder stage.  \cite{zhu2022visually} designs a two-stage pipeline, where the second stage conducts a counterfactual synthesis through motion features to remove potentially leaked sound.  The approach of \cite{gao2021visualvoice} added critic functions explicitly to enforce cross-modal consistency and cross-instrument contrast.

\paragraph{Vision Transformers.}
Motivated by Transformer's success in natural language processing~\cite{vaswani2017attention}, transformers are first introduced in computer vision for image classification as ViT~\cite{dosovitskiy2020image}. Given the superior long-range modeling capacity, many follow-up works~\cite{touvron2021training,yuan2021tokens,mao2021dual} have upgraded ViT to achieve higher performance and widely surpassed convolutional neural networks. Further, transformer-based models are adopted for various downstream tasks, such as 2D object detection~\cite{zhu2020deformable,carion2020end,zheng2020end}, semantic/instance segmentation~\cite{strudel2021segmenter,zheng2021rethinking,xie2021segformer}, 3D object detection~\cite{zhang2022monodetr,misra2021end}, shape recognition~\cite{zhang2022point,zhao2021point} and video understanding~\cite{arnab2021vivit,liu2022video}.
Particularly, following the pipeline from DETR~\cite{carion2020end}, MaskFormer~\cite{cheng2021per} and Mask2Former~\cite{cheng2021masked} represent each mask candidate as a learnable query and conduct parallel decoding for instance-level segmentation. However, only few approaches \cite{rahman2021tribert,zhu2022visually,lee2021looking,tzinis2020into,tzinis2022audioscopev2} have extended Transformer for visual-audio sound separation fields. \cite{rahman2021tribert} adopts a BERT \cite{devlin2018bert} architecture to learn visual, pose, and audio feature representations. \cite{zhu2022visually} designs an Audio-Motion Transformer to refine sound separation results through motion-audio feature fusion. These methods focus mainly on learning better contextualized multi-modality representations through an encoder Transformer. In contrast, our Mask Transformer-based network focuses on the entire process of visual-audio separation task. We disentangle different instruments, namely, sound sources, through independent learnable query prototypes and segment each time-frequency region on the spectrum via mask prediction in an end-to-end fashion. 


\section{Method}

We first describe the formulation of the visual sound separation task and introduce our pipeline iQuery briefly in Sec. \ref{3.1}. Then we introduce networks for learning representations from visual and audio modalities in Sec. \ref{3.2} and our proposed cross-modality cross-attention transformer architecture for visual sound separation in Sec. \ref{3.3}. Finally, we introduce our adaptive query fine-tuning strategy through designs of flexible tunable queries in Sec. \ref{3.4}. 

\subsection{Overview \label{3.1}}
As mentioned before, our goal is to disentangle audio mixture concerning its corresponding sound sources in the given mixture through the usage of so-called queries. Following previous works \cite{ephrat2018looking,zhao2018sound}, we adopt a commonly used 
``Mix-and-Separate" self-supervised source separation procedure. Given $K$ video clips with accompanying audio signal: $\{(V_i, s_i(t))\}_{i \in [1, K]}$, we create a sound mixture: $s_{mix}(t) = \sum_{i=1}^K s_i(t)$ as training data. Our disentanglement goal is to separate sounds $s_i(t)$ from $s_{mix}(t)$ for sound sources in $V_i$, respectively.
The pipeline, as illustrated in Fig. \ref{fig:method}, is mainly composed of two components: an Audio-Visual Feature Extraction module and a Mask Transformer-based Sound Separation module. First, in the feature extraction module, the object detector \& image encoder, and video encoder extract object-level visual features and motion features from video clip $V_i$. The audio network yields an audio feature and an audio embedding from the given sound mixture $s_{mix}(t)$. After that, a cross-modal transformer decoder attends to visual and audio features and outputs audio mask embeddings, which would further be combined with audio embeddings for separate sound prediction.

\subsection{Audio-Visual Feature Extraction \label{3.2}}


\paragraph{Object Detection Network \& Image Encoder.}
To initialize learning of audio queries, we assign object-level visual appearance features to the corresponding queries, to create a ``visually named'' query. In the implementation, following \cite{gao2019co,tian2021cyclic}, we use a Faster R-CNN object detector with ResNet-101 backbone trained on Open Images \cite{krasin2017openimages}. For frames in a given video clip $V_i$, the pre-trained object detector is utilized to acquire the detected objects set $O_i$. After that, we adopt a pre-trained ResNet-18 similar to \cite{tian2021cyclic}, followed by a linear layer and max pooling to yield object-level features $F_{O_i} \in \mathbb{R}^{C_O} $, where $C_O$ denotes channel dimension of object features.

\paragraph{Video Encoder.}
The video encoder maps the video frame from $V_i \in \mathbb{R}^{3 \times T_i \times H_i \times W_i}$ into a motion feature representation. In contrast with previous motion representations \cite{zhao2019sound,gan2020music,rahman2021tribert,zhu2022visually}, we use self-supervised video representation obtained from a 3D video encoder of I3D \cite{carreira2017quo} pre-trained by FAME \cite{ding2022motion}. The model is pre-trained contrastively to concentrate on moving foregrounds. Finally, a spatial pooling is applied to obtain motion embedding $F_{M_i} \in \mathbb{R}^{C_M \times T_i'}$, where $C_M$ denotes the dimension of the motion feature. 

\paragraph{Audio Network.}
The audio network takes the form of skip-connected U-Net style architectures \cite{ronneberger2015u} following \cite{zhao2018sound,gao2019co,tian2021cyclic}. Given the input audio mixture \emph{$s_{mix}(t)$}, we first apply a Short-Time Fourier Transform (STFT) \cite{griffin1984signal} to convert the raw waveform to a 2D Time-Frequency spectrogram representation $S_{mix} \in \mathbb{R}^{F \times T}$, which is then fed into the U-Net encoder to obtain an audio feature map \emph{$F_A \in \mathbb{R}^{C_A \times \frac{F}{S} \times \frac{T}{S}}$}($C_A$ denotes the number of channels and $S$ denotes stride of audio feature map) at the bottleneck. A U-Net decoder gradually upsamples the audio features to yield audio embeddings \emph{$\varepsilon_A \in \mathbb{R}^{C_\varepsilon \times F \times T}$} ($C_\varepsilon$ denotes the dimension of audio embeddings), which would be combined further with the Transformer mask embeddings to generate the separated sound spectrogram mask \emph{$M_i$}.


\subsection{Audio-Visual Transformer \label{3.3}}
Our cross-modality sound separation transformer contains the Transformer decoder \cite{vaswani2017attention} with $N$ queries (\textit{i}.\textit{e}., learnable prototypes), and utilizes the extracted object features $F_{O_i}$, motion embeddings $F_{M_i}$ and audio features $F_A$ to yield $N$ mask embeddings $\varepsilon_{mask} \in \mathbb{R}^{C_{\varepsilon} \times N}$ for spectrogram mask prediction of separated sound $s_i(t)$, where $N$ denotes maximum of the pre-defined instrument types following \cite{cheng2021per}. 

\paragraph{Audio query prototypes.}
We denote audio queries as $Q \in \mathbb{R}^{C_Q \times N}$ to represent different instruments, which are initialized by ``visually naming" sound queries. Specifically, ``visually naming'' stands for that we assign object features $F_{O_i}$ to corresponding query $Q_i$ with element-wise addition to yield ``visually-named" queries $Q_v$, which would be then fed into the Transformer decoder cross-attention layers. 

\paragraph{Cross-attention layers.}
In the decoder, we stack one motion-aware decoder layer and three audio-aware decoder layers. The ``visually-named'' queries $Q_v$ would first interact temporally with extracted motion features $F_{M_i}$ in the motion-aware decoder layer with motion cross-attention by $\mathrm{Attention}(Q_v, F_{M_i}, F_{M_i})$.
This is followed by an FFN to generate the motion-decoded query $Q'$, which would be fed into three following audio-aware decoder layers to adaptively interact with audio features $F_A$, each of which consists of a self-attention, an audio cross-attention computed by $\mathrm{Attention}(Q', F_A, F_A)$, and an FFN. Thus, the output $N$ audio segmentation embeddings $\varepsilon_Q \in \mathbb{R}^{C_Q \times N}$ is computed by
\begin{equation}
      \varepsilon_Q = \mathrm{AudioDecoder}_{\times 3}(Q',F_A,F_A),
\end{equation}
where $\mathrm{AudioDecoder}$ stands for our audio-aware decoder layer. Similar to \cite{carion2020end,cheng2021per}, the decoder generates all audio segmentation embeddings parallelly. 

\paragraph{Separated mask prediction.}
Through the above decoder, the $N$ audio segmentation embeddings $\varepsilon_Q$  are converted to $N$ mask embeddings $\varepsilon_{mask} \in \mathbb{R}^{C_{\varepsilon} \times N}$ through a MLP with two hidden layers, where dimension $C_{\varepsilon}$ is identical to dimension of audio embeddings \emph{$\varepsilon_A \in \mathbb{R}^{C_\varepsilon \times F \times T}$}. Then each predicted mask $M_i \in \mathbb{R}^{F \times T}$ of the separated sound spectrogram is generated by a dot-product between the corresponding mask embedding $\varepsilon_{mask_{i}}$ and audio embedding $\varepsilon_A$ from the audio decoder.
Finally, we multiply the sound mixture spectrogram $S_{mix}$ and the predicted mask $M_i$ to disentangle sound spectrogram $S_i$ for sound $s_i(t)$ by
\begin{equation}
        S_i = S_{mix} \odot M_i,
\end{equation}
where $\odot$ denotes the element-wise multiplication operator. Ultimately, separated sound waveform signal $s_i(t)$ is produced by applying inverse STFT to the separated spectrogram $S_i$.

\paragraph{Training objective.}
Following \cite{zhao2018sound,gao2019co}, we set our training objective as optimizing spectrogram masks. The ground truth ratio mask $M_i^{GT}$ of $i$-th video is calculated as follows,
\begin{equation}
        M_i^{GT} (t,f) = \frac{S_i(t,f)}{S_{mix}(t,f)},
\end{equation}
where $(t,f)$ denotes time-frequency coordinates. We adopt per-pixel $L1$ loss \cite{zhao2016loss} to optimize the overall sound separation network, sound separation loss $L_{sep}$ is defined as,
\begin{equation}
        L_{sep} = \sum_{i=1}^K ||M_i - M_i^{GT} ||_1,
\end{equation}
where $K$ denotes number of mixed sounds in $S_{mix}$.

\subsection{Tunable Queries as Audio Prompts  \label{3.4}}

\begin{figure}[t]
\begin{center}
    \includegraphics[width=\linewidth]{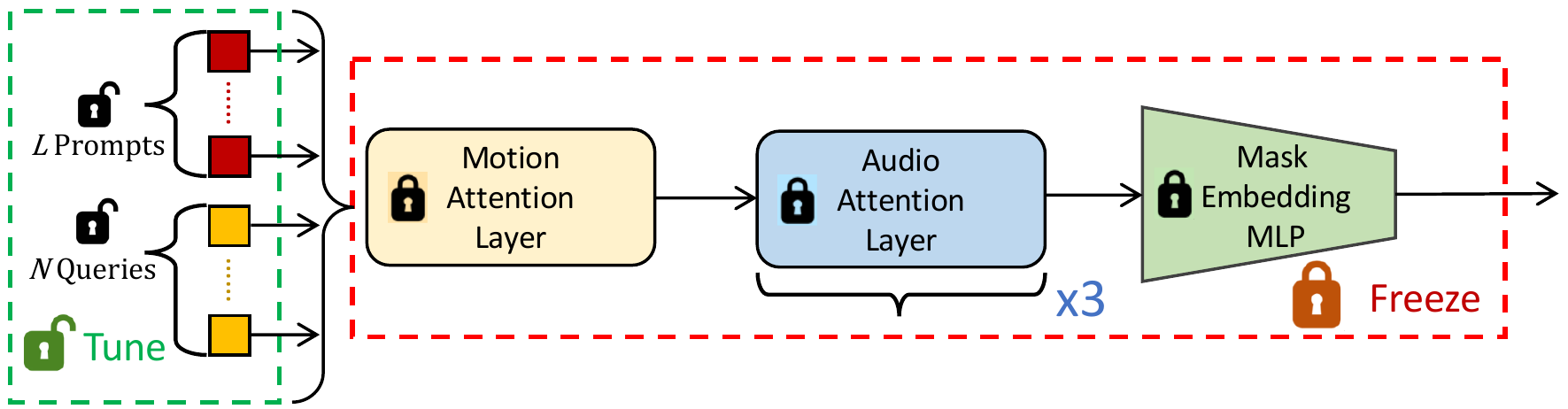}
\end{center}
\vspace{-0.2cm}
\caption{\textbf{Audio prompts design.} When comes to generalizing to new types of instruments/event classes, we propose to insert additional queries (\emph{audio prompts}) to learn new audio prototypes for unseen classes. With this design, we only tune the query embedding layer while keeping all the other parts of Transformer backbone frozen for fine-tuning.}
\label{fig:audio prompt}
\end{figure}

With the flexible design of tunable queries as learnable prototypes, our pipeline is more friendly to generalizing new types of instruments. Unlike previous methods that need to finetune the entire mask generation U-Net, we could insert additional queries(\textit{i}.\textit{e}., audio prompts) for the new instruments. Such a method enables us only need to finetune the query embedding layer for learning new audio query prototypes in Sec. \ref{3.3} of our Transformer architecture while keeping all cross-attention layers frozen (see Fig.\ref{fig:audio prompt}). Specifically, we add a new audio prompt $P \in \mathbb{R}^{C_Q}$ to original pre-trained audio queries $q \in \mathbb{R}^{C_Q \times N}$, then the query embedding layer for the prompted learnable prototypes $q_{prompted} \in \mathbb{R}^{C_Q \times (N+1)}$ is the only layer learnable in our Transformer decoder, while keeping the Transformer backbone frozen.

\section{Experiments}
\subsection{Experimental Settings}
\paragraph{Datasets.} We perform experiments on three widely-used datasets: \emph{MUSIC} \cite{zhao2018sound}, \emph{MUCIS-21} \cite{zhao2019sound}, and \emph{Audio-Visual Event (AVE)} \cite{gemmeke2017audio,tian2018audio}. \emph{MUSIC} dataset spans 11 musical instrument categories: accordion, acoustic guitar, cello, clarinet, erhu, flute, saxophone, trumpet, tuba, violin, and xylophone. This dataset is relatively clean, and sound sources are always within the scene, collected for the audio-visual sound separation task. We utilize 503 online available solo videos and split them into training/validation/testing sets with 453/25/25 videos from 11 different categories, respectively, following the data and splits settings of \cite{tian2021cyclic}. \emph{MUSIC-21} dataset \cite{zhao2019sound} is an enlarged version of \emph{MUSIC} \cite{zhao2018sound}, which contains 10 more common instrument categories: bagpipe, banjo, bassoon, congas, drum, electric bass, guzheng, piano, pipa, and ukulele. We utilize 1,092 available solo videos and split them into training/testing sets with 894/198 videos respectively from 21 different categories. Note that we follow the same training/testing split as \cite{gan2020music,zhu2022visually}. \emph{AVE} dataset is a general audio-visual learning dataset, covering 28 event classes such as animal behaviors, vehicles, and human activities. We follow the same setting as \cite{zhu2022visually}, and utilize 4143 videos from AVE \cite{tian2018audio} dataset. 

\paragraph{Baselines.} For \emph{MUSIC} dataset, we compare our method with four recent methods for sound separation. \emph{NMF-MFCC} \cite{spiertz2009source} is a non-learnable audio-only method, we consider reporting this result from \cite{gao2019co,rahman2021tribert} on \emph{MUSIC} test set. We also compare our method with two representative audio-visual sound separation baselines: \emph{Sound-of-Pixels} \cite{zhao2018sound} and \emph{CO-SEPARATION} \cite{gao2019co}. We retrained these two methods with the same training data and split them as ours for a fair comparison. Finally, we compare our approach with a most recent publicly-available baseline \emph{CCoL} \cite{tian2021cyclic}, which has the same training setting as ours.
For \emph{MUSIC-21} dataset, we compare our method with six recently proposed approaches: \emph{Sound-of-Pixels} \cite{zhao2018sound}, \emph{CO-SEPARATION}\cite{gao2019co}, \emph{Sound-of-Motions} \cite{zhao2019sound}, \emph{Music Gesture} \cite{gan2020music}, \emph{TriBERT} \cite{rahman2021tribert} and \emph{AMnet} \cite{zhu2022visually} on MUSIC-21 dataset \cite{zhao2019sound}. For \cite{rahman2021tribert}, since $12.27 \% $ of the training samples are missing in their given training split, we consider their reported result as a baseline comparison. 
Finally, for \emph{AVE} dataset, we compare our method with six state-of-the-art methods. Since we conduct our experiments with the same setting as \emph{AMnet} \cite{zhu2022visually}, we report results from \cite{zhu2022visually} for \emph{Multisensory} \cite{owens2018audio}, \emph{Sound-of-Pixels} \cite{zhao2018sound}, \emph{Sound-of-Motions} \cite{zhao2019sound}, \emph{Minus-Plus} \cite{xu2019recursive}, \emph{Cascaded Opponent Filter} \cite{zhu2020visually} as baseline comparisons.



\paragraph{Evaluation metrics.} The sound separation performance is evaluated by the popular adopted mir\_eval library \cite{raffel2014mir_eval} in terms of standard metrics: Signal to Distortion Ratio (SDR), Signal to Interference Ratio (SIR), and Signal to Artifact Ratio (SAR). SDR and SIR scores measure the separation accuracy, and SAR captures the absence of artifacts. For all three metrics, a higher value indicates better results.

\begin{figure}
\begin{center}
    \includegraphics[width=0.5\linewidth]{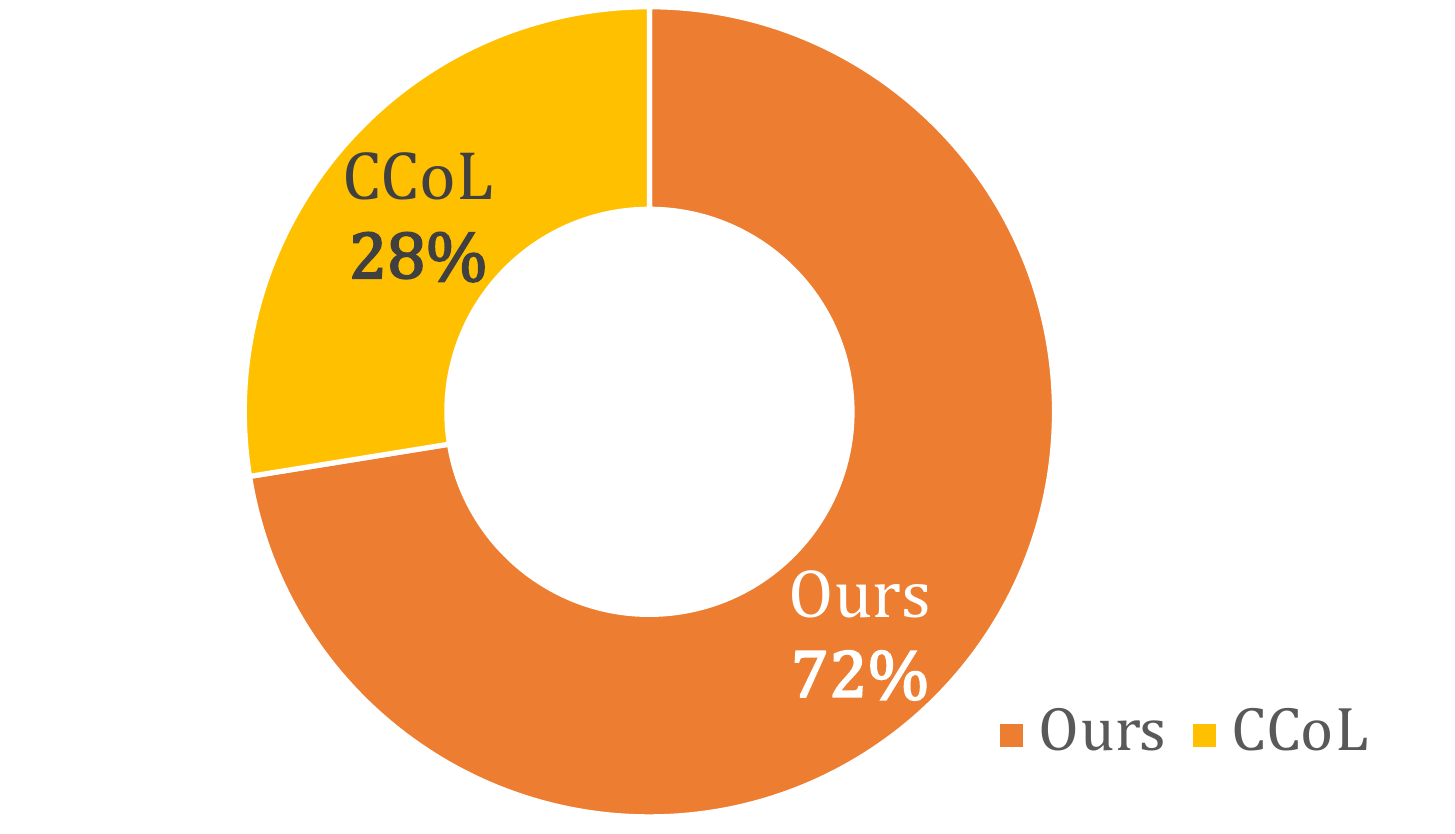}
\end{center}
\caption{\textbf{Human evaluation results} for sound source separation on mixtures of different instrument types. Our system is able to separate sounds with better actual perceptual quality.}
\label{fig:human}
\end{figure}
\begin{figure}[t]
\begin{center}
    \includegraphics[width=0.8\linewidth]{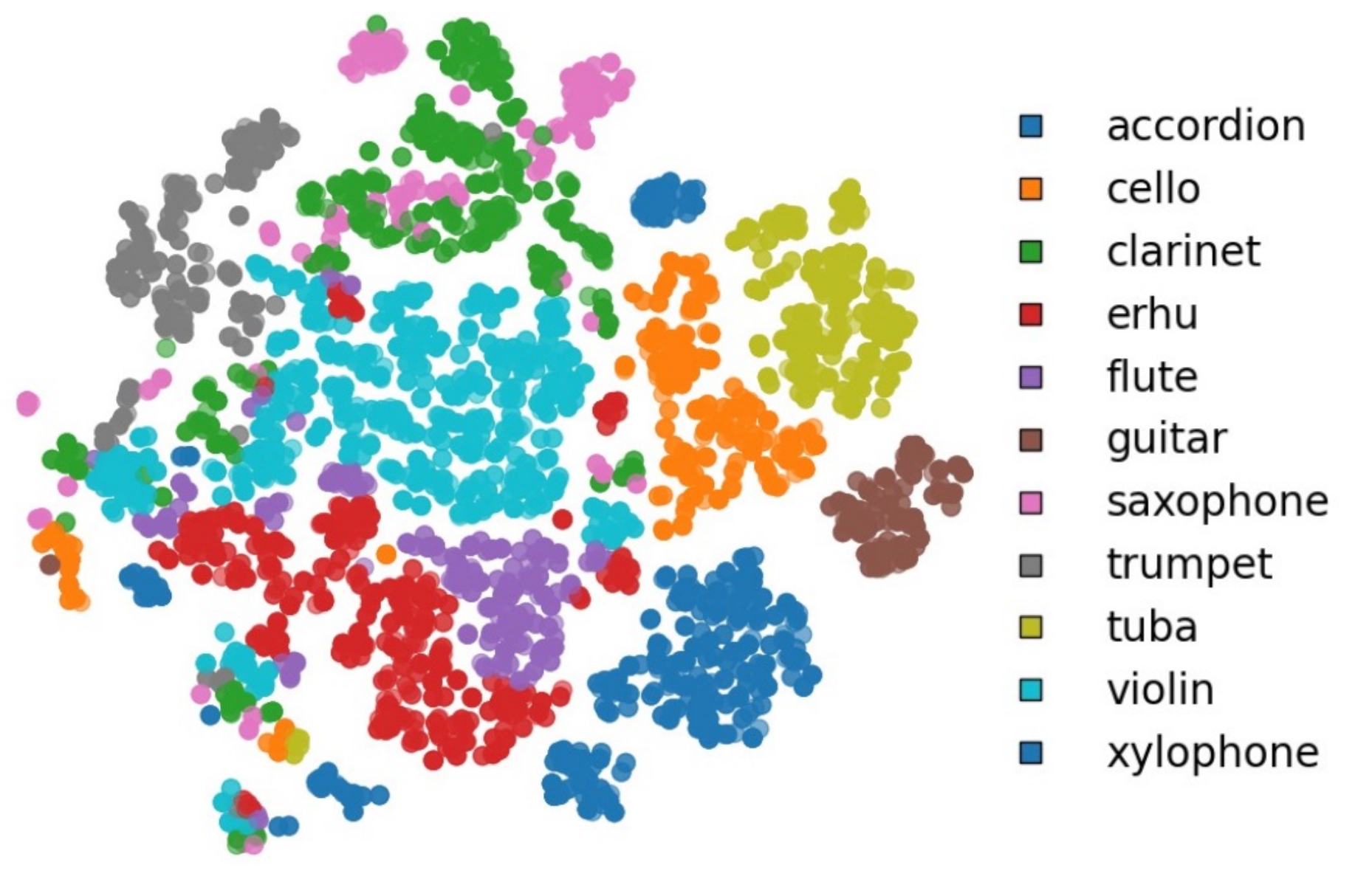}
\end{center}
\vspace{-2mm}
\caption{\textbf{Visualization of audio query embeddings} with t-SNE, different instrument categories are color-coded. Our audio queries have learned to cluster by different classes of sound.}
\label{fig:tsne}
\end{figure}

\paragraph{Implementation Details.} For \emph{MUSIC} \cite{zhao2018sound} and \emph{MUSIC-21} \cite{zhao2019sound} datasets, we sub-sample the audio at 11kHz, and each audio sample is approximately 6 seconds. STFT is applied using a Hann window size of 1022 and a hop length of 256, yielding a 512 $\times$ 256 Time-Frequency audio representation. It is then re-sampled on a log-frequency scale to obtain a magnitude spectrogram with $T,F =$ 256. Detected objects in frames are resized to 256$\times$256 and randomly cropped to the size of 224$\times$224. We set the video frame rate as $1fps$, and randomly-selected three frames are inputted into the object detector. While for \emph{AVE} \cite{tian2018audio} dataset, audio signal is sub-sampled at 22kHz, and we use the full frame rate ($29.97fps$). Other settings are the same as \emph{MUSIC} except STFT hop length is set as 184, following \cite{zhu2022visually}.

For \emph{MUSIC} dataset \cite{zhao2018sound}, we use the Faster R-CNN object detector pre-trained by \cite{gao2019co} on Open Images \cite{krasin2017openimages}.
For \emph{MUSIC-21} \cite{zhao2019sound} and \emph{AVE} \cite{tian2018audio} datasets, since additional musical and general classes are not covered for this object detector, we adopt a pre-trained Detic detector \cite{zhou2022detecting} based on CLIP \cite{radford2021learning} to detect the 10 more instruments in \emph{MUSIC-21} dataset \cite{zhao2019sound} and 28 event classes in \emph{AVE} dataset \cite{tian2018audio}.

We utilize 8 heads for all attention modules and select the maximum $N$ objects (number of queries) as 15/30. The vide encoder \cite{ding2022motion} and the object detection network is pre-trained and kept frozen during training and inference. The multi-layer perception (MLP) for separated mask prediction has 2 hidden layers of 256 channels following \cite{cheng2021per}. Audio feature $F_A$, motion feature $F_M$, object feature $F_O$, and audio queries $Q$ have a channel dimension of 256. And we set the channel dimension of both audio embeddings $\varepsilon_A$ and mask embeddings $\varepsilon_M$ as 32. The epoch number is 80, and batch size is set to 8. We use AdamW \cite{loshchilov2017decoupled} for the mask transformer with a weight decay of $10^{-4}$ and Adam for all other networks as optimizer selection. The learning rate of the transformer is set as $10^{-4}$ and will decrease by multiplying 0.1 at 60-th epoch. We set the learning rate for other networks as $10^{-4}$, decreased by multiplying 0.1 at 30-th and 50-th epoch, respectively. Training is conducted on 8 NVIDIA TITAN V GPUs. 

\begin{figure}[t]
\begin{center}
\vspace{10mm}
    \includegraphics[width=0.7\linewidth]{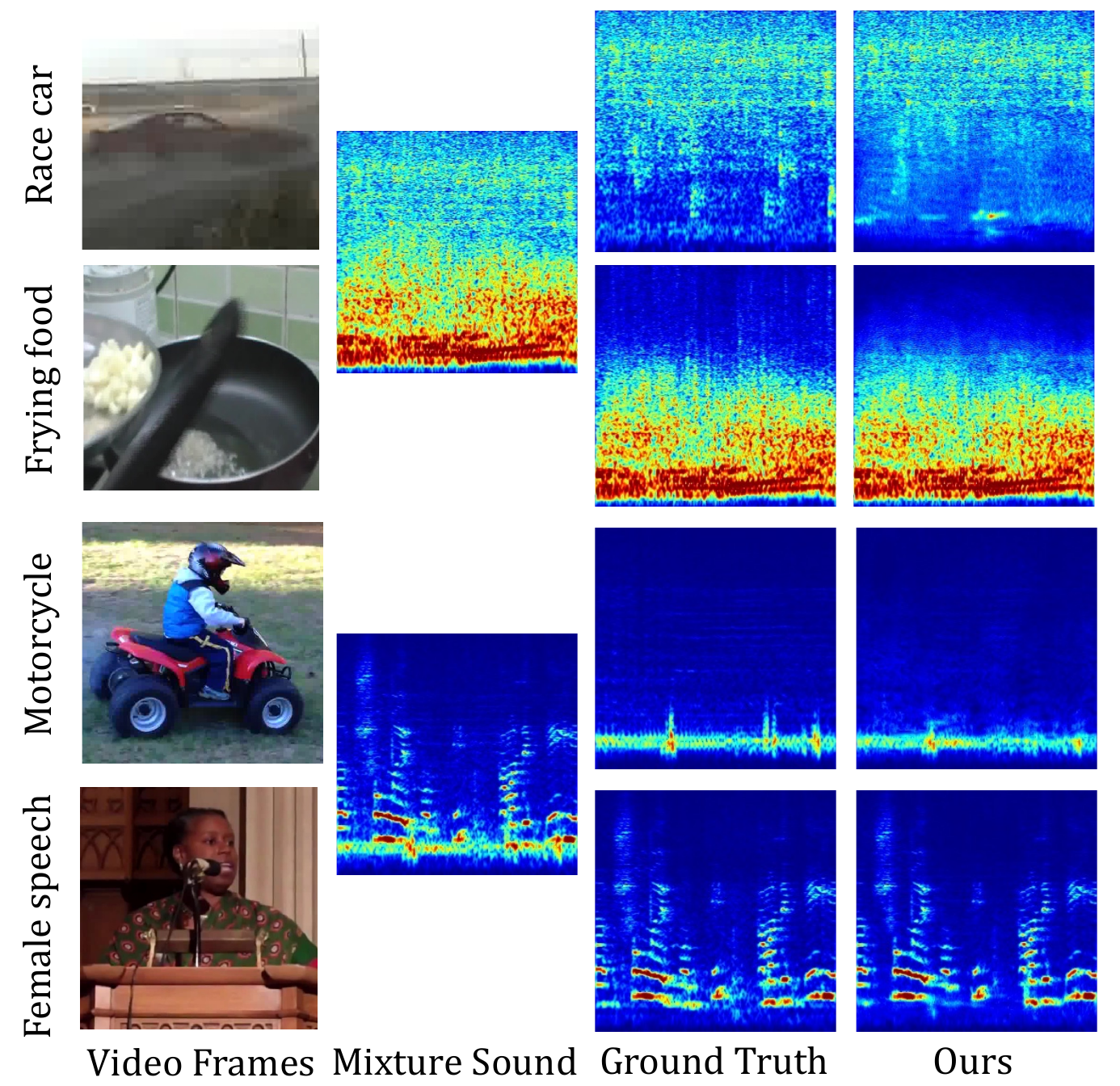}
\end{center}
\vspace{-2mm}
\caption{\textbf{Qualitative results on AVE test dataset.} Beyond restricted musical instruments, our model is also able to handle general sound separation tasks (\textbf{e.g.} sounds of galloping race car and frying food on the first two rows; sounds of driving motorcycles and speeches on the last two rows).}
\label{fig:avequali}
\end{figure}

\subsection{Audio-Visual Sound Source Separation}
\paragraph{Quantitative evaluation.} Table. \ref{tab:quanta} demonstrates quantitative results for sound separation results against state-of-the-art methods on \emph{MUSIC} dataset \cite{zhao2018sound}. Our method outperforms baseline models in separation accuracy measured by all evaluation metrics. Our method outperforms the most recent publicly available state-of-the-art algorithm \cite{tian2021cyclic} by 3.43 dB in terms of SDR score. Regarding quantitative results on \emph{MUSIC21} dataset \cite{zhao2019sound}, we demonstrate the performance comparison in Table. \ref{tab:music21}. Again, our method outperforms baseline models in terms of SDR metric. 
Performance on the previous two datasets demonstrate our model's ability to disentangle musical sounds. To further verify the scalability of our proposed method to general audio-source separation problems, we perform quantitative comparisons on \emph{AVE} dataset in Table. \ref{tab:ave}. As is demonstrated, we surpass the state-of-the-art algorithm \cite{zhu2022visually} by 1.31 dB in terms of SDR score. \emph{AVE} is a general dataset containing scenes like male\&female speeches, animal sounds, and vehicle sounds. This clearly shows our model's adaptivity to more general problems of sound source separation.

\paragraph{Qualitative evaluation.} Fig. \ref{fig:qualitative} further illustrates qualitative sound separation results on \emph{MUSIC} dataset. It can be seen that our method disentangles sound sources cleaner and more accurately, with less ``muddy" sound. Fig. \ref{fig:avequali} provides additional qualitative examples on \emph{AVE} dataset, and this again illustrates our model's good performance on general sound source separation cases. Both qualitative and quantitative results verify the superiority of our designed sound query-based segmentation pipeline iQuery. 


\begin{table}[tb]
\centering
\begin{tabular}{l|ccc}
\toprule
Methods & SDR$\uparrow$ & SIR$\uparrow$ & SAR$\uparrow$ \\
\midrule
NMF-MFCC \cite{spiertz2009source} &0.92 & 5.68 & 6.84\\
Sound-of-Pixels \cite{zhao2018sound} & 4.23 & 9.39 & 9.85 \\
CO-SEPARATION \cite{gao2019co} & 6.54 & 11.37 & 9.46 \\
CCoL \cite{tian2021cyclic} & \textcolor{blue}{7.74} & \textcolor{blue}{13.22} & \textcolor{blue}{11.54} \\
\midrule
\textbf{iQuery (Ours)} & \textbf{11.17} & \textbf{15.84} & \textbf{14.27} \\
\bottomrule
\end{tabular}
\vspace{-2mm}
\caption{\textbf{Audio-visual sound separation results on \emph{MUSIC}.} Best results in \textbf{bold} and second-best results in \textcolor{blue}{Blue}.}
\label{tab:quanta}
\end{table}
\begin{table}[t]
\centering
\begin{tabular}{l|ccc}
\toprule
Methods & SDR$\uparrow$ & SIR$\uparrow$ & SAR$\uparrow$ \\
\midrule
Sound-of-Pixels \cite{zhao2018sound}* &7.52 & 13.01 & 11.53\\
CO-SEPARATION \cite{gao2019co}* &7.64 & 13.80 & 11.30\\
Sound-of-Motions \cite{zhao2019sound}* & 8.31 & 14.82 & 13.11 \\
Music Gesture \cite{gan2020music}* & 10.12 & 15.81 & - \\
TriBERT \cite{rahman2021tribert} & 10.09 & \textcolor{blue}{17.45} & 12.80 \\
AMnet \cite{zhu2022visually}* & \textcolor{blue}{11.08} & \textbf{18.00} & \textcolor{blue}{13.22} \\
\midrule
\textbf{iQuery (Ours)} & \textbf{11.12} & 15.98 & \textbf{14.16} \\
\bottomrule
\end{tabular}
\vspace{-2mm}
\caption{\textbf{Audio-visual sound separation results on \emph{MUSIC-21}.} The results noted by * are obtained from \cite{gan2020music,zhu2022visually}.}
\label{tab:music21}
\end{table}
\paragraph{Human evaluation.} Our quantitative evaluation shows the superiority of our model compared with baseline models, however, studies \cite{cano2016evaluation} have shown that audio separation quality could not be truthfully determined purely by the widely used mir\_eval \cite{raffel2014mir_eval} metrics. Due to this reason, we further conduct a subjective human evaluation to study the actual perceptual quality of sound-separation results.
Specifically, we compare the sound separation result of our model and the publicly available best baseline model \cite{tian2021cyclic} on \emph{MUSIC} \cite{zhao2018sound}. We collected 50 testing samples for all 11 classes from the test set, and each testing sample contains separated sounds with a length of 6 seconds predicted by our model and baseline \cite{tian2021cyclic} for the same sound mixture. Ground truth sound is also provided for each sample as a reference. The experiment is conducted by 40 participants separately. For each participant, the orders of our model and baseline \cite{tian2021cyclic} are randomly shuffled, and we ask the participant to answer ``\emph{Which sound separation result is more close to the ground truth audio?}" for each sample. Statistical results are shown in Fig. \ref{fig:human}. Notably, our proposed method significantly surpasses the compared baseline with a winning rate of 72.45\%. This additionally demonstrate the better actual perceptual performance of our model.

\paragraph{Learned Query Embedding.} To visualize that our proposed model has indeed learned to disentangle different sound sources through learnable queries, we display a t-SNE embeddings (see Fig. \ref{fig:tsne}) of our learnable queries in MUSIC \cite{zhao2018sound} test set. It can be seen that our queries tend to cluster by different instrument categories, learning representative prototypes.

\begin{table}[tb]
\centering
\begin{tabular}{l|ccc}
\toprule
Methods & SDR$\uparrow$ & SIR$\uparrow$ & SAR$\uparrow$ \\
\midrule
Multisensory \cite{owens2018audio}* &0.84 & 3.44 & 6.69\\
Sound-of-Pixels \cite{zhao2018sound}* &1.21 & 7.08 & 6.84\\
Sound-of-Motions \cite{zhao2019sound}* & 1.48 & 7.41 & 7.39 \\
Minus-Plus \cite{xu2019recursive}* & 1.96 & 7.95 & 8.08 \\
Cascaded Filter \cite{zhu2020visually}* & 2.68 & 8.18 & 8.48 \\
AMnet \cite{zhu2022visually}* & \textcolor{blue}{3.71} & \textbf{9.15} & \textcolor{blue}{11.00} \\
\midrule
\textbf{iQuery (Ours)} & \textbf{5.02} & \textcolor{blue}{8.21} & \textbf{12.32} \\
\bottomrule
\end{tabular}
\vspace{-2mm}
\caption{\textbf{Audio-visual sound separation results on \emph{AVE}.}  The results noted by * are obtained from \cite{zhu2022visually}.}
\label{tab:ave}
\end{table}

\begin{table}[tb]
\centering
\begin{tabular}{l|ccc}
\toprule
Methods & SDR$\uparrow$ & SIR$\uparrow$ & SAR$\uparrow$ \\
\midrule
Sound-of-Pixels \cite{zhao2018sound} &4.11 & 8.17 & 9.84\\
CO-SEPARATION \cite{gao2019co} & 5.37 & 9.85 & 8.72 \\
CCoL \cite{tian2021cyclic} & \textcolor{blue}{6.74} & \textbf{11.94} & \textcolor{blue}{10.22} \\
\midrule
\textbf{iQuery (Ours)}& \textbf{8.04} & \textcolor{blue}{11.60} & \textbf{13.21} \\
\bottomrule
\end{tabular}
\vspace{-2mm}
\caption{\textbf{Fine-tuning sound separation performance comparison.} All methods are pretrained on \emph{MUSIC} dataset without one particular instrument and then fine-tuned on this new data. Baseline models are tuned with whole network unfrozen, and we keep our transformer backbone frozen. }
\label{tab:fine}
\end{table}

\begin{figure}[t]
    \centering
    \subfloat[]{
    \begin{minipage}{0.49\linewidth}
        \centering
        \includegraphics[width=\linewidth]{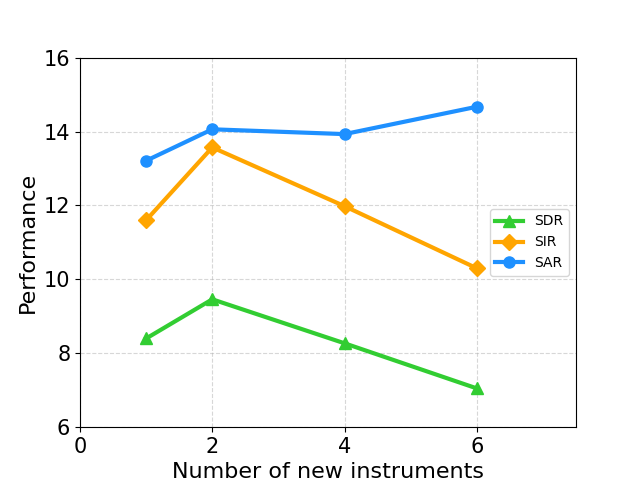}
    \end{minipage}}
    \subfloat[]{
    \begin{minipage}{0.49\linewidth}
        \centering
        \includegraphics[width=\linewidth]{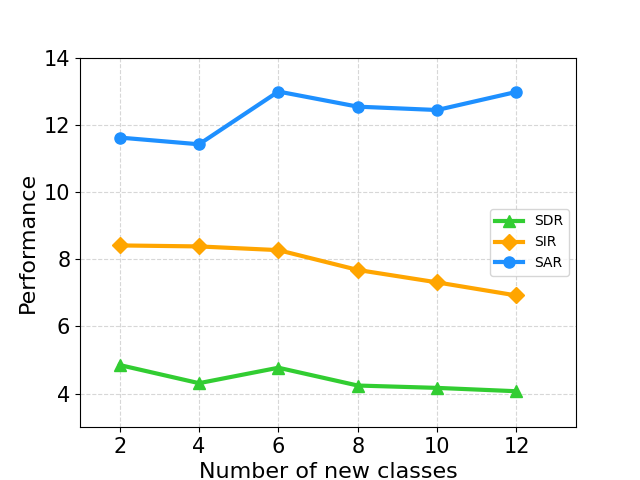}
    \end{minipage}}
    \vspace{-2mm}
    \caption{\textbf{Fine-tuning curves} of sound separation. (a) Fine-tuning with different number of unseen instrument classes on \emph{MUSIC}. (b) Fine-tuning with different number of unseen event classes on \emph{AVE}.}
    \label{fig:finecurves}
\end{figure}

\subsection{Extenable Audio Prompt Fine-tuning}

Table. \ref{tab:fine} evaluates our approach's generalization ability compared with previous methods. We conduct fine-tuning experiments by leave-one-out cross-validation. 
Baseline models are fine-tuned on the new instrument with all the networks structure unfrozen.  With the design of audio prompts discussed in Sec. \ref{3.4}, we keep most of our transformer parameters frozen,  only fine-tuning the query embedding layer, which has much fewer parameters (0.048\% of the total parameters in Transformer).

Fig. \ref{fig:finecurves} (a) shows our performance with a varying number of new instrument classes for fine-tuning on \emph{MUSIC} dataset. We hold out 1, 2, 4, and 6 instrument classes in the pre-training stage and fine-tune our method on these new classes with only the query embedding layer unfrozen. \emph{MUSIC} dataset contains in total of 11 instruments. Notably, our method still yields good results when the network is only pre-trained on 5 instrument types, even fewer than the unseen classes. Fig. \ref{fig:finecurves} (b) shows our model's fine-tuning performance on \emph{AVE} dataset with a varying number of new event classes for fine-tuning. We follow the experimental setup on \emph{MUSIC}, and hold out 2, 4, 6, 8, and 12 event classes for fine-tuning. This demonstrates our model's adaptivity in general sound separation cases.

\subsection{Contrastive Verification \label{4.4}}
Our learnable query-prototypes network is designed to ensure cross-modality consistency and cross-instrument contrast. We assume these query-prototypes to draw samples of each particular sound class sample close and push away the different prototypes. The question is whether our network design with ``visually-named'' query trained in the ``Mix-and-Separate'' manner can already achieve this goal? As an alternative,  we design an auxiliary contrastive loss for verification: to maximize the cosine similarity of separated audio embedding $\varepsilon_{A_i} = \varepsilon_A \odot M_i$ and the corresponding query embedding $Q_i$, while minimizing the cosine similarity of separated audio embedding and other query embeddings ${Q_n}$ (where ${n \in [1,N], n \neq i}$).   We optimize the cross-entropy losses of these cosine similarity scores to obtain our contrastive loss $L_{contras}$.   To ensure the qualities of audio embedding $\varepsilon_A$ and predicted mask $M_i$ is accurate enough, we take a Hierarchical Task Learning strategy \cite{lu2021geometry} to control weights for $L_{sep}$ and $L_{contras}$ at each epoch. The verification loss $L_{verify}$ is:
$
        L_{verify} = w_{sep}(t) \cdot L_{sep} + w_{contras}(t) \cdot L_{contras}
$
where $t$ denotes current training epoch and $w(t)$ denotes loss weight. 

\paragraph{Ablations of auxiliary contrastive loss}, shown in Table. \ref{tab:abl2}, demonstrates that our existing design achieves better results without using explicit contrastive loss. This answers the question we raised, that our ``visually-named" queries are already contrastive enough for sound disentanglement.

\begin{table}[tb]
\centering
\small
\begin{tabular}{l|ccc}
\toprule
Architecture & SDR$\uparrow$ & SIR$\uparrow$ & SAR$\uparrow$ \\
\midrule
w/o lrn.  & 10.05 & 14.27 & 13.71 \\
w/o adpt.   & 10.89 & 15.51 & 14.14 \\
w/ con. best & 11.02 & 15.91 & 14.10 \\
\midrule
Ours (w/o con) &11.17 & 15.84 & 14.27 \\
\bottomrule
\end{tabular}
\vspace{-2mm}
\caption{\textbf{Ablations on the auxiliary contrastive loss.} ``w/o lrn." denotes without learnable linear layer added to queries produced by Transformer decoder; ``w/o adpt." denotes that we use a fixed weight for auxiliary contrastive loss without the Hierarchical Task Learning strategy; ``w/ con. best" denotes our best model design using auxiliary contrastive loss.}
\label{tab:abl2}
\end{table}

\begin{table}[tb]
\centering
\small
\begin{tabular}{l|ccc}
\toprule
Architecture & SDR$\uparrow$ & SIR$\uparrow$ & SAR$\uparrow$ \\
\midrule
Random & 6.58 & 10.79 & 12.77 \\
Self-audio & 10.54 & 14.81 & 14.23 \\
Self-motion-audio  & 10.65 & 15.37 & 13.96 \\
Dual-stream   & 10.46 & 15.25 & 13.79 \\
\midrule
Motion-self-audio &11.17 & 15.84 & 14.27 \\
\bottomrule
\end{tabular}
\vspace{-2mm}
\caption{\textbf{Ablations on the design of Transformer decoder.}}
\label{tab:abl1}
\end{table}

\subsection{Ablations of Transformer decoder design}
Ablation results of Transformer decoder design is shown in Table. \ref{tab:abl1}. ``Random" denotes randomly assigning object features to queries; its poor separation result verifies the importance of our ``visually-named" queries. ``Self-audio" means removing the motion cross attention layer, which confirms the effectiveness of adding the motion feature. We tried two baseline designs against our final selection ``motion-self-audio'', as stated in Sec. \ref{3.3}. ``Self-motion-audio" is a design that puts self-, motion cross-, and audio cross-attention in a single decoder layer. ``Dual-stream" means we conduct motion and audio cross-attention in parallel then fuse in the decoder layer. Specific details are in the Supplemental material.










\section{Conclusion}
We developed an audio-visual separation method using an adaptable query-based audio Mask Transformer network. Our network disentangles different sound sources explicitly through learnable audio prototypes initiated by the ``visually named'' audio queries.   
We demonstrate cross-modal consistency and cross-instrument contrast via a multi-modal cross-attention mechanism. When generalizing to a new unseen instrument, our method can be adapted by inserting an additional query as an audio prompt while freezing the attention mechanism. Experiments on both musical and general sound datasets demonstrate performance gain by our iQuery.

\paragraph{Limitation and Discussion.} The paradigm of ``Mix-and-Separate" framework enables us to use current public datasets for training, might somewhat limit our model's applicability to real mixtures. If sound sources are not visible in the scene, we plan to explore our system's ability in on/off-screen separation tasks when dataset of \cite{tzinis2020into,tzinis2022audioscopev2} is available. Our pipeline is capable of extracting sound sources on the same hierarchy level (``dog" vs.``piano"), and we will explore sound separation from a general category of ``animals" and ``music".  Finally, while we mainly focus on the problem of Audio-Visual Sound Separation, our concept of attention with audio queries could also be applied to other tasks like visual sound source localization and recently proposed audio-visual segmentation \cite{zhou2022audio}.



{\small
\bibliographystyle{ieee_fullname}
\bibliography{egbib}
}

\end{document}